\documentclass[runningheads]{llncs}
\usepackage{graphicx}

\usepackage{tikz}
\usepackage{comment}
\usepackage{amsmath,amssymb} 
\usepackage{color}

\usepackage[accsupp]{axessibility}  
\usepackage{boldline}
\usepackage{float}
\usepackage{bm}
\usepackage{bbm}
\usepackage{booktabs}
\usepackage{tablefootnote}
\usepackage{pifont}
\newcommand{\cmark}{\ding{51}}%
\newcommand{\xmark}{\ding{55}}%
\newcommand{\myparagraph}[1]{\medbreak\noindent\textbf{#1}}

\begin{document}
\pagestyle{headings}
\mainmatter
\def\ECCVSubNumber{3}  

\title{STC: Spatio-Temporal Contrastive Learning for Video Instance Segmentation} 

\titlerunning{Spatio-Temporal Contrastive Learning for Video Instance Segmentation}
%
\author{Zhengkai Jiang\inst{1}\thanks{Equal contributions. This work was done while Zhangxuan Gu was interning at Tencent Youtu Lab.} \and 
Zhangxuan Gu\inst{2*} \and Jinlong Peng\inst{1} \and
Hang Zhou\inst{3} \and \\ Liang Liu\inst{1} \and Yabiao Wang\inst{1} \and Ying Tai \inst{1} \and Chengjie Wang\inst{1} \and Liqing Zhang\inst{2}}
\authorrunning{Zhengkai Jiang et al.}
%
\institute{Tencent Youtu Lab \and
Shanghai Jiao Tong University \and The Chinese University of Hong Kong\\
\email{\{zhengkjiang, caseywang\}@tencent.com}}
\maketitle

\begin{abstract}
Video Instance Segmentation (VIS) is a task that simultaneously requires classification, segmentation, and instance association in a video. Recent VIS approaches rely on sophisticated pipelines to achieve this goal, including RoI-related operations or 3D convolutions. In contrast, we present a simple and efficient single-stage VIS framework based on the instance segmentation method CondInst by adding an extra tracking head. To improve instance association accuracy, a novel bi-directional spatio-temporal contrastive learning strategy for tracking embedding across frames is proposed. Moreover, an instance-wise temporal consistency scheme is utilized to produce temporally coherent results. Experiments conducted on the YouTube-VIS-2019, YouTube-VIS-2021, and OVIS-2021 datasets validate the effectiveness and efficiency of the proposed method. We hope the proposed framework can serve as a simple and strong baseline for other instance-level video association tasks.
\end{abstract}

\section{Introduction}
\label{sec:intro}
While significant progress~\cite{he2017mask,bolya2019yolact,wang2020centermask,liu2018path,xie2020polarmask,wang2020solo,wang2020solov2,tian2020conditional,lin2021video,oksuz2021rank} has been made in instance segmentation with the development of deep neural networks, less attention has been paid to its challenging variant in the video domain. The video instance segmentation (VIS)~\cite{yang2019video,yang2021crossover,wang2021end,hwang2021video} task requires not only classifying and segmenting instances but also capturing the instance associations across frames. Such technology can benefit a great variety of scenarios, {\em e.g.,} video editing, video surveillance, autonomous driving, and augmented reality. As a result, it is in great need of accurate, robust, and fast video instance segmentation approach in practice.

\begin{figure}
\centering
\includegraphics[width=0.49\textwidth]{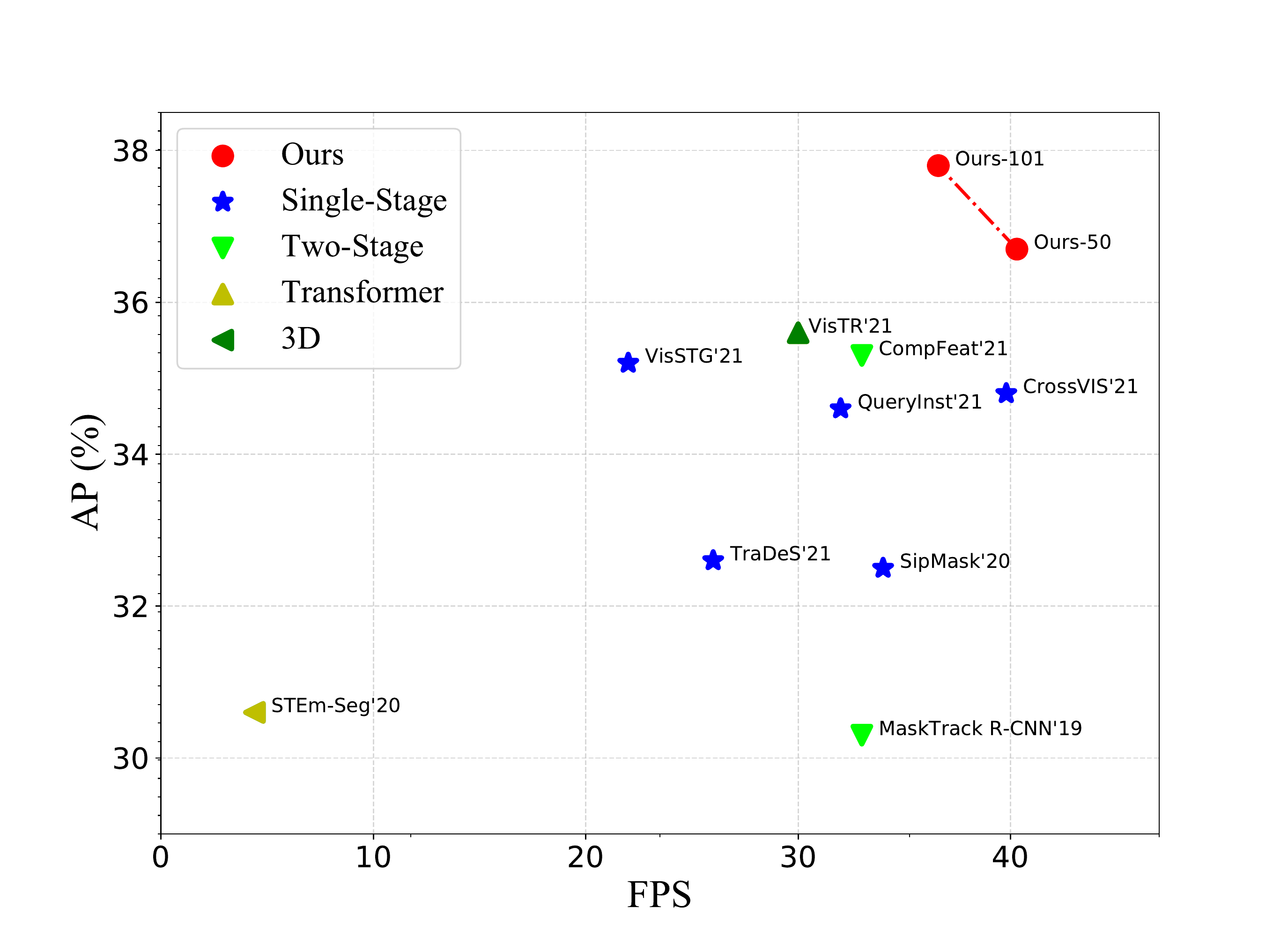}
\caption{Speed-Accuracy trade-off curve on the YouTube-VIS-2019 validation set. The baseline results are compared with the same ResNet-50 backbone for fair comparison. We achieve best tradeoff between speed and accuracy. In particular, STC exceeds recent CrossVIS~\cite{fang2021instances} 1.9\% mAP with similar running speed.}
\label{FPS_AP}
\end{figure}

Previous researchers have developed sophisticated pipelines for tackling this problem~\cite{yang2018efficient,voigtlaender2019feelvos,yang2019video,cao2020sipmask,bertasius2020classifying,wang2021end,athar2020stem,fu2020compfeat}. 
Generally speaking, previous studies can be divided into the categories of two-stage~\cite{yang2018efficient,voigtlaender2019feelvos,yang2019video,bertasius2020classifying,fu2020compfeat}, feature-aggregation~\cite{liu2021sg,bertasius2020classifying} inspired from video object detection domain~\cite{zhu2017flow,jiang2020learning,jiang2019video}, 3D convolution-based~\cite{athar2020stem}, transformer-based~\cite{wang2021end,hwang2021video}, and single-stage ~\cite{cao2020sipmask,yang2021crossover} methods. Two-stage methods, \emph{e.g.}, MaskTrack R-CNN~\cite{yang2019video} and CompFeat~\cite{fu2020compfeat}, usually rely on the RoIAlign operation to crop the feature and obtain the representation of an instance for further binary mask prediction. 
Such the RoIAlign operation would lead to great computational inefficiency. 3D convolution-based STEm-Seg~\cite{athar2020stem} holds huge complexity and could not achieve good performance. Transformer-based VisTR~\cite{wang2021end,hwang2021video} could not handle long videos due to largely increasing memory usage and needs a much longer training time for convergence. Feature-aggregation methods~\cite{li2021spatial,wangtao2021end} enhance features through pixel-wise or instance-wise aggregation from adjacent frames similarly to other video tasks, like video object detection~\cite{jiang2020learning,jiang2019video,wu2019sequence}. Although some attempts~\cite{cao2020sipmask,wu2021track,yang2021crossover} have been made to tackle VIS in a simple single-stage manner, their performances are still not satisfying.

The key difference between video and image instance segmentation lies in the need of capturing robust and accurate instance association across frames. However, most previous works such as MaskTrack R-CNN~\cite{yang2019video}, and CMaskTrack R-CNN~\cite{qi2021occluded} formulate instance association as a multi-label classification problem, focusing only on the intrinsic relationship within instances while ignoring the extrinsic constraint between different ones. Thus different instances with similar distributions may be wrongly associated by using previous tracking embeddings only through such multi-label classification loss constraint. 

Alternatively, we propose an \textit{efficient} single-stage fully convolutional network for video instance segmentation task, considering that single-stage instance segmentation is simpler and faster. Based on the recent instance segmentation method CondInst~\cite{tian2020conditional}, an extra tracking head is added to simultaneously learn instance-wise tracking embeddings for instance association besides original classification head, box head, and mask head by dynamic filter. To improve instance association accuracy between adjacent frames, a spatio-temporal contrastive learning strategy is utilized to exploit relations between different instances. Specifically, for a tracking embedding query, we densely sample hundreds of negative and positive embeddings from reference frames based on the label assignment results, acting as a contrastive manner to jointly pull closer to the same instances and push away from different instances. Different from previous metric learning based instance association methods {\em i.e., Triplet Loss}, the proposed contrastive strategy enables efficient many-to-many relations learning across frames. We believe this contrast mechanism enhances the instance similarity learning, which provides more substantial supervision than using only the labels. Moreover, this contrastive learning scheme is applied in a bi-directional way to better leverage the temporal information from both forward and backward views. At last, we further propose a temporal consistency scheme for instance encoding, which contributes to both the accuracy and smoothness of the video instance segmentation task. 

In summary, our main contributions are: 
\begin{itemize}
    \item We propose a single-stage fully convolutional network for video instance segmentation task with an extra tracking head to simultaneously generate instance-specific tracking embeddings for instance association.
    \item To achieve accurate and robust instance association, we propose a bi-directional spatio-temporal contrastive learning strategy that aims to obtain representative and discriminative tracking embeddings. In addition, we present a novel temporal consistency scheme for instances encoding to achieve temporally coherent results.
    \item Comprehensive experiments are conducted on the YouTube-VIS-2019, YouTube-VIS-2021, and OVIS-2021 benchmark. Without bells and whistles, we achieve 36.7\% AP and 35.5\% AP with ResNet-50 backbone on YouTube-VIS-2019 and YouTube-VIS-2021 datasets, which is the best performance among all listed single-model methods with high efficiency. We also achieve best performance on recent proposed OVIS-2021 dataset. In particular, compared to the first VIS method named MaskTrack R-CNN~\cite{yang2019video}, our proposed method (STC) achieves 36.7\% AP on YouTube-VIS-2019, outperforming it by \textbf{6.4}\% AP with the advantage of being much simpler and faster. Compared with recent method CrossVIS~\cite{yang2021crossover}, STC outperforms it by 1.9\% AP with a slightly faster speed. 
\end{itemize}
\section{Related Works}
\subsection{Instance Segmentation}
Instance segmentation aims to represent objects at a pixel level, which is a finer-grained representation compared with object detection. There are mainly two kinds of instance segmentation methods, {\em i.e.,} two-stage~\cite{he2017mask,liu2018path,huang2019mask}, and single-stage~\cite{chen2020blendmask,bolya2019yolact,wang2020solo,wang2020solov2,tian2020conditional}. Two-stage methods first detect objects, then crop their region features to further classify each pixel into the foreground or background, while the framework of single-stage instance segmentation is much simpler. For example, YOLACT~\cite{bolya2019yolact} is proposed to generate a set of prototype masks and predict per-instance mask coefficients. The instance masks are then produced by linearly combining the prototypes with the mask coefficients. SOLO~\cite{wang2020solo,wang2020solov2} reformulates the instance segmentation as two simultaneous category-aware prediction problems, \emph{i.e.}, location prediction, and mask prediction, respectively. Inspired by dynamic filter network~\cite{jia2016dynamic}, CondInst~\cite{tian2020conditional} proposes to dynamically predict instance-aware filters for mask generation. SOLOv2~\cite{wang2020solov2} further incorporates dynamic filter scheme to dynamically segments each instance in the image with a novel matrix non-maximum suppression (NMS) technique. Compared to the image instance segmentation, video instance segmentation aims not only to segment object instances in individual frames but also to associate the predicted instances across frames.

\begin{figure*}[t]
\centering
\includegraphics[width=0.95\textwidth]{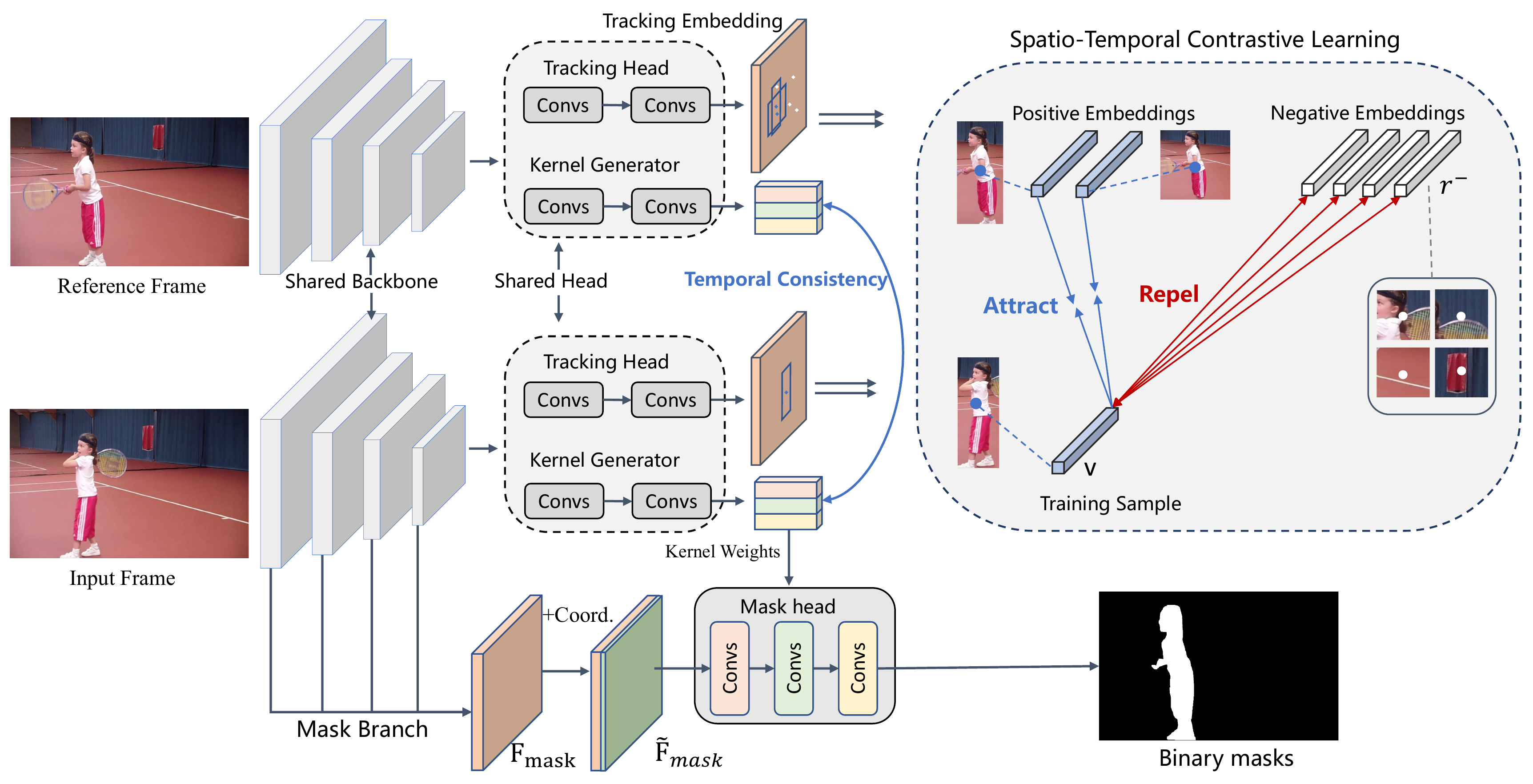}
\caption{\textbf{The overview of our proposed framework.} The framework contains the following components: a shared CNN backbone for encoding frames to feature maps, kernel generators with mask heads for instance segmentation, a mask branch to combine multi-scale FPN features, and a shared tracking head with a bi-directional spatio-temporal contrastive learning strategy (the bi-directional learning scheme is omitted here for simplicity) for instance association. A temporal consistency constraint is applied to the kernel weights, as the blue line shows. Best viewed in color.}
\label{framework}
\end{figure*}

\subsection{Video Instance Segmentation}
Video instance segmentation~\cite{yang2019video} aims to simultaneously classify, segment, and track instances of the videos. Various complicated pipelines are designed by state-of-the-art methods to solve it. To better introduce the related methods, we separate them into the following groups. (1) The two-stage method MaskTrack R-CNN~\cite{yang2019video}, as the pioneering work for VIS, extends image instance segmentation method Mask R-CNN~\cite{he2017mask} to video domain by introducing an extra tracking branch for instance association. Another method in the two-stage group is MaskProp~\cite{bertasius2020classifying}, which first uses Hybrid Task Cascade (HTC)~\cite{chen2019hybrid} to generate the predicted masks and propagates them temporally to the other frames in a video. Recently, CompFeat~\cite{fu2020compfeat} proposed a feature aggregation approach based on MaskTrack R-CNN, which refines features by aggregating multiple adjacent frames features. (2) Relying on 3D convolutions, STEm-Seg~\cite{athar2020stem} models a video clip as a single 3D spatial-temporal volume and separates object instances by clustering. (3) Based on feature-aggregation, STMask~\cite{li2021spatial} proposes a simple spatial feature calibration to detect and segment object masks frame-by-frame, and further introduces a temporal fusion module to track instances across frames. (4) More recently, a transformer-based method VisTR~\cite{wang2021end} is proposed to reformulate VIS as a parallel sequence decoding problem. (5) There also exist some single-stage VIS methods, {\em e.g.,} SipMask~\cite{cao2020sipmask}, and TraDeS~\cite{wu2021track}. SipMask~\cite{cao2020sipmask} proposes a spatial preservation module to generate spatial coefficients for the mask predictions while recently proposed TraDeS~\cite{wu2021track} presents a joint detection and tracking model by propagating the previous instance features with the predicted tracking offset. CrossVIS~\cite{yang2021crossover} proposes cross-frame instance-wise consistency loss for video instance segmentation. Although current methods have made good progress, their complicated pipelines or unsatisfying performance prohibit practical application. In contrast, the proposed framework acts in a fully convolutional manner with decent performance and efficiency.

\subsection{Contrastive Learning}
Contrastive learning has lead to considerable progress in many real-world applications~\cite{he2020momentum,chen2020simple,tian2020makes,xiong2020loco,pang2021quasi,kalantidis2020hard,chen2020improved}. For example, MOCO~\cite{he2020momentum} builds image-level large dictionaries for unsupervised representation learning using contrastive loss. SimCLR~\cite{chen2020simple} utilizes the elaborate data augmentation strategies and a large batch, which outperforms MOCO by a large margin on self-supervised learning ImageNet~\cite{russakovsky2015imagenet} classification task. 
Different from the above methods, which focus on image-level contrastive learning for unsupervised representation learning, we use modified multiple-positives contrastive learning to learn instance-level tracking embeddings accurately for video instance segmentation tasks. 
\section{Method}
In this section, we first briefly review the instance segmentation method CondInst~\cite{tian2020conditional} for mask generation of still-image. Then, we introduce the proposed whole framework for the video instance segmentation task. Next, we present a novel spatio-temporal contrastive learning strategy for tracking embeddings to achieve accurate and robust instance association. In addition, we further propose a bi-directional spatio-temporal contrastive learning strategy. At last, the temporal consistency scheme aiming to achieve temporally coherent results is introduced in detail. 

\subsection{Mask Generation for Still-image}
For still-image instance segmentation, we use the dynamic conditional convolutions method CondInst~\cite{tian2020conditional,jia2016dynamic}. Specifically, instance mask at location $(x, y)$ can be generated by convolving an instance-agnostic feature map $\mathbf{\tilde{F}}_{mask}^{x, y}$ from mask branch and instance-specific dynamic filter $\bm{\theta}_{x,y}$, which is calculated as follows:
\begin{equation}
    \mathbf{m}_{x,y} = \mathbf{MaskHead}(\mathbf{\tilde{F}}_{mask}^{x,y};\bm{\theta}_{x,y}),
\end{equation}
where $\mathbf{\tilde{F}}_{mask}^{x, y}$ is the combination of multi-scale fused feature map $\mathbf{{F}}_{mask}$ from FPN features $\{P_3, P_4, P_5\}$ and relative coordinates $\mathbf{O}_{x,y}$. The $\mathbf{MaskHead}$ consists of three $1\times1$ conv-layers with dynamic filter $\bm{\theta}_{x,y}$ at location $(x, y)$ as convolution kernels. $\mathbf{m}_{x,y} \in \mathbb{R}^{H\times W}$ is the predicted binary mask at location $(x,y)$ as shown in Figure~\ref{framework}.

\subsection{Proposed Framework for VIS}
The overall framework of the proposed method is illustrated in Figure~\ref{framework}. Based on the instance segmentation method CondInst~\cite{tian2020conditional}, we add a tracking head for instances association. The whole architecture mainly contains following components: (1) A shared CNN backbone (\emph{e.g.} ResNet-50~\cite{he2016deep}) is utilized to extract compact visual feature representations with FPN~\cite{lin2017feature}. (2) Multiple heads including a classification head, a box regression head, a centerness head, a kernel generator head, and a mask head as same as CondInst~\cite{tian2020conditional}. Since the architectures of the above classification, box regression, and centerness heads are not our main concerns, we omit them here (please refer to \cite{tian2019fcos} for the details). (3) A tracking head where spatio-temporal contrastive learning strategy is proposed to associate instances across frames with comprehensive relational cues in the tracking embeddings. (4) Temporal consistency scheme on instance-wise kernel weights across frames aims to generate temporally coherent results.

\subsection{Spatio-Temporal Contrastive Learning}
To associate instances from different frames, an extra lightweight tracking head is added to obtain the tracking embeddings~\cite{yang2019video,cao2020sipmask,fu2020compfeat} in parallel with the original kernel generator head. The tracking head consists of several convolutional layers which take multi-scale FPN features $\{P_3, P_4, P_5\}$ as input. And the outputs are fused to obtain the feature map of tracking embedding. As shown in Figure~\ref{framework}, given an input frame $I$ for training, we randomly select a reference frame $I_{ref}$ from its temporal neighborhood. A location is defined as a positive sample if it falls into any ground-truth box and the class label $c$ of the location is the class label of the ground-truth box. If a location falls into multiple bounding boxes, it is considered as the positive sample of the bounding box with minimal area~\cite{tian2019fcos}. Thus, two locations formulate a positive pair if they are associated with the same instance across two frames and a negative pair otherwise.

During training, for a given frame, the model first predicts the object detection results. Then, the tracking embedding of each instance can be extracted from the tracking feature map by the center of the predicted bounding box. For a training sample with extracted tracking embedding $q$, we can obtain positive embeddings $\mathbf{k}^+$ and negative embeddings $\mathbf{k}^-$ according to label assignment results at reference frame. Note that traditional  unsupervised representation learning~\cite{he2020momentum,chen2020simple} with contrastive learning only uses one positive sample and multiple negative samples as follows: 

\begin{equation}
\label{contrast_eqn}
\begin{aligned}
\mathcal{L}_{q} = - \log\frac{\exp(\mathbf{q} \cdot \mathbf{k}^+)}{\exp(\mathbf{q} \cdot \mathbf{k}^+)+\sum_{\mathbf{k}^-}\exp(\mathbf{q}\cdot \mathbf{k}^-)}.
\end{aligned}   
\end{equation}

Since there are many positive embeddings at reference frame for each training sample, instead of randomly selecting one positive embedding at reference frames, we optimize the objective loss with multiple positive embeddings and multiple negative embeddings as:
\begin{equation}
\label{track_eqn}
\begin{aligned}
\mathcal{L}_{contra} &= - \sum_{\mathbf{k}^+}\log\frac{\exp(\mathbf{q} \cdot \mathbf{k}^+)}{\exp(\mathbf{q} \cdot \mathbf{k}^+)+\sum_{\mathbf{k}^-}\exp(\mathbf{q}\cdot \mathbf{k}^-)} \\
&= \sum_{\mathbf{k}^+} \log [1 + \sum_{\mathbf{k}^-}\exp(\mathbf{q} \cdot \mathbf{k}^- - \mathbf{q} \cdot \mathbf{k}^+)].
\end{aligned}   
\end{equation}
Suppose there are $N_{pos}$ training samples at input frame, the objective track loss with multiple samples is:

\begin{equation}
\label{track_loss}
\begin{aligned}
\mathcal{L}_{track} = \frac{1}{N_{pos}} \sum_{i=1}^{N_{pos}} \mathcal{L}_{contra}^{i}.
\end{aligned}   
\end{equation}

\noindent \textbf{Bi-directional Spatio-Temporal Learning.} Many video-related tasks have shown the effectiveness of bi-directional modeling~\cite{zhu2017bidirectional,sun2019learning}. To fully exploit such temporal context information, we further propose a bi-directional spatio-temporal learning scheme to learn instance-wise tracking embeddings better. Note that we only utilize this scheme in the training stage, and thus it does not affect the inference speed. Similar to Equation~\ref{track_loss}, the objective function of bi-directional spatio-temporal contrastive learning can be denoted as $\mathcal{\hat{L}}_{track}$ by reversing input frame and reference frame. Thus, the final bi-directional spatio-temporal contrastive loss is:
\begin{equation}
\mathcal{L}_{bi-track}= \frac{1}{2} (\mathcal{L}_{track} + \mathcal{\hat{L}}_{track}).
\end{equation}

\subsection{Temporal Consistency}
Compared with image data, the coherent property between frames is also crucial to video-related researches. Thus, we add a temporal consistency constraint on the kernel weights, marked as the blue line in Figure~\ref{framework}, to capture such prior during training so that the predicted masks will be more accurate and robust across frames. Given an instance at location $(x, y)$ appearing at both input and reference frames, we use $(x, y)$ and $(\hat{x}, \hat{y})$ to denote its positive candidate positions from two frames, respectively. 
Formally, the temporal consistency constraint during training can be formulated as an L2-loss function:
\begin{equation}
    \mathcal{L}_{consistency} =||\bm{\theta}_{x,y} - \bm{\theta}_{\hat{x},\hat{y}}^{ref}||^2 + ||\bm{m}_{x,y} - \bm{m}_{\hat{x},\hat{y}}^{ref}||^2,
\end{equation}
where $\bm{\theta}_{\hat{x},\hat{y}}^{ref}$ is the dynamic filter at reference frame, $\bm{m}_{\hat{x},\hat{y}}^{ref}$ is the predicted instance mask by reference dynamic filter. With such a simple constraint, our kernel generator can obtain accurate, robust and coherent mask predictions across frames.

\subsection{Training and Inference} 

\noindent \textbf{Training Scheme.} Formally, the overall loss function of our model can be formulated as follows:
\begin{equation}
\mathcal{L}_{overall}= \mathcal{L}_{condinst}+ \lambda_b \mathcal{L}_{bi-track} + \lambda_c \mathcal{L}_{consistency},
\end{equation}
where $\mathcal{L}_{condinst}$ denotes the original loss of CondInst~\cite{tian2020conditional} for instance segmentation. We refer readers to \cite{tian2020conditional} for the details of $\mathcal{L}_{condinst}$. $\lambda_b$ and $\lambda_c$ are the hyper-parameters.

\noindent \textbf{Inference on Frame.} For each frame, we forward it through the model to get the outputs, including classification confidence, centerness scores, box predictions, kernel weights, and tracking embeddings. Then we obtain the box detections by selecting the positive positions whose classification confidence is larger than a threshold (set as 0.03), similar to FCOS~\cite{tian2019fcos}. After that, following previous work MaskTrack R-CNN~\cite{yang2019video}, the NMS~\cite{bolya2019yolact} with the threshold being 0.5 is used to remove duplicated detections. In this step, these boxes are also associated with the kernel weights and tracking embeddings. Supposing that there remain $T$ boxes after the NMS, thus we have $T$ groups of the generated kernel weights. Then $T$ groups of kernel weights are used to produce $T$ mask heads. These instance-specific mask heads are applied to the positions encoded mask feature to predict the instance masks following~\cite{tian2020conditional}. $T$ is 10 in default following previous work MaskTrack R-CNN.

\begin{table*}[t!]
\caption{Comparisons with some state-of-the-art approaches on \textbf{YouTube-VIS-2019} $\mathtt{val}$ set. \checkmark indicates using extra data augmentation ({\em e.g.}, random crop, higher resolution input, multi-scale training) ~\cite{bertasius2020classifying} or additional data \cite{athar2020stem,bertasius2020classifying,fu2020compfeat,wang2021end}. $^{\dagger}$ indicates the method that reaches higher performance by stacking multiple networks, and we regard it an unfair competitor in general setting. 
 Note that STMask~\cite{li2021spatial} uses deformable convolution network (DCN)~\cite{dai2017deformable} as the backbone, which is still inferior to our method at both accuracy and speed, demonstrating the superiority of our proposed framework. $^{\dagger\dagger}$ means transformer on top of ResNet-50 or ResNet-101.}
\centering
\resizebox{\textwidth}{!}{
\small
 \begin{tabular}{l|c|c|cc|ccccc}
 \hlineB{2}
  Method 
  & Publication
  & Augmentations
  & Backbone
  & FPS
  & AP 
  & AP$_{\mathtt{50}}$ 
  & AP$_{\mathtt{75}}$ 
  & AR$_{\mathtt{1}}$ 
  & AR$_{\mathtt{10}}$ \\
  \hline
  MaskTrack R-CNN~\cite{yang2019video} & ICCV'19 & \xmark & ResNet-50 & 33 & 30.3 & 51.1 & 32.6 & 31.0 & 35.5 \\
  SipMask~\cite{cao2020sipmask} & ECCV'20 & \xmark & ResNet-50 & 34 & 32.5 & 53.0 & 33.3 & 33.5 & 38.9 \\
  STEm-Seg~\cite{athar2020stem}  & ECCV'20 & \xmark & ResNet-50 & 4.4 & 30.6 & 50.7 & 33.5 & 31.6 & 37.1 \\
  CompFeat~\cite{fu2020compfeat}  & AAAI'21 & \xmark & ResNet-50 &  $<$ 33 & 35.3 & 56.0 & 38.6 & 33.1 & 40.3  \\
  TraDeS~\cite{wu2021track} & CVPR'21 & \xmark & ResNet-50 & 26 & 32.6 & 52.6 & 32.8 & 29.1 & 36.6 \\
  QueryInst~\cite{fang2021instances} & ICCV'21 & \xmark & ResNet-50 & 32 & 34.6 & 55.8 & 36.5 & 35.4 & 42.4 \\
  CrossVIS~\cite{yang2021crossover} & ICCV'21 & \xmark & ResNet-50 & 39.8 & 34.8 & 54.6 & 37.9 & 34.0 & 39.0 \\
  VisSTG~\cite{wangtao2021end} & ICCV'21 & \xmark & ResNet-50 & 22 & 35.2 & 55.7 & 38.0 & 33.6 & 38.5 \\
  PCAN~\cite{ke2021prototypical} & NeurIPS'21 & \xmark & ResNet-50 & - & 36.1 & 54.9 & \bfseries39.4 & 36.3 & 41.6 \\
  \bfseries Ours (STC) & - & \xmark & ResNet-$50$ & \bfseries40.3 & \bfseries36.7 & \bfseries57.2 & 38.6 & \bfseries36.9 & \bfseries44.5    \\
  \hlineB{1}
  STMask~\cite{li2021spatial} & CVPR'21 & DCN backbone~\cite{dai2017deformable} & ResNet-50 & 29 & 33.5 & 52.1 & 36.9 &
31.1 & 39.2 \\
  SG-Net~\cite{liu2021sg} & CVPR'21 & multi-scale training & ResNet-50 & 23 & 34.8 & 56.1 & 36.8 & 35.8 & 40.8 \\
  VisTR~\cite{wang2021end} & CVPR'21 & random-crop training & ResNet-50 & 30 & 35.6 & 56.8 & 37.0 & 35.2 & 40.2 \\ 
  QueryInst~\cite{fang2021instances} & ICCV'21 & multi-scale training & ResNet-50 & 32 & 36.2 & 56.7 & 39.7 & 36.1 & 42.9 \\
  CrossVIS~\cite{yang2021crossover} & ICCV'21 & multi-scale training & ResNet-50 & 39.8 & 36.3 & 56.8 & 38.9 & 35.6 & 40.7 \\
  VisSTG~\cite{wangtao2021end} & ICCV'21 & multi-scale training & ResNet-50 & 22 & 36.5 & 58.6 & 39.0 & 35.5 & 40.8 \\
  \bfseries Ours (STC) & - & multi-scale training & ResNet-$50$ & \bfseries40.3 & \bfseries37.6 & \bfseries58.9 & \bfseries39.7 & \bfseries38.2 & \bfseries46.2    \\
  \hlineB{1}
  MaskTrack R-CNN~\cite{yang2019video} & ICCV'19 & \xmark & ResNet-101 & 33 & 30.3 & 51.1 & 32.6 & 31.0 & 35.5 \\
  SRNet~\cite{ying2021srnet} & ACMMM'21 & \xmark & ResNet-101 & 35 & 32.3 & 50.2 & 34.8 & 32.3 & 40.1 \\
  STEm-Seg~\cite{athar2020stem}  & ECCV'20 & \xmark & ResNet-101 & 2.1 & 34.6 & 55.8 & 37.9 & 34.4 & 41.6 \\
  PCAN~\cite{ke2021prototypical} & NeurIPS'21 & \xmark & ResNet-101 & - & 37.6 & 57.2 & \bfseries41.3 & 37.2 & 43.9 \\
  \bfseries Ours (STC) & - & \xmark & ResNet-$101$ & \bfseries36.6 & \bfseries37.8 & \bfseries58.5 & 40.6 & \bfseries38.5 & \bfseries46.3    \\
  \hlineB{1}
  SipMask~\cite{cao2020sipmask} & ECCV'20 & multi-scale training & ResNet-101 & 24 & 35.8 & 56.0 & 39.0 & 35.4 & 42.4 \\
  STMask~\cite{li2021spatial} & CVPR'21 & DCN backbone~\cite{dai2017deformable} & ResNet-101 & 23 & 36.8 & 56.8 & 38.0 &
34.8 & 41.8 \\
  SG-Net~\cite{liu2021sg} & CVPR'21 & multi-scale training & ResNet-101 & 20 & 36.3 & 57.1 & 39.6 & 35.9 & 43.0 \\
  VisTR~\cite{wang2021end} & CVPR'21 & random-crop training & ResNet-101 & 28 & 38.6 & 61.3 & 42.3 & 37.6 & 44.2 \\ 
  \bfseries Ours (STC) & - & multi-scale training & ResNet-$101$ & \bfseries36.6 & \bfseries39.2 & \bfseries61.5 & \bfseries42.4 & \bfseries39.7 & \bfseries47.3    \\
  \hlineB{2}
 \end{tabular}}
 \label{sota}
\end{table*}

\noindent \textbf{Inference on Video.} Given a testing video, we first construct an empty memory bank for the predicted instance embeddings. Then our model processes each frame sequentially in an online scheme. Our network generates a set of predicted instance embeddings at each frame. The association with identified instances from previous frames relies on the cues of embedding similarity, box overlap, and category label similar to the MaskTrack R-CNN~\cite{yang2019video}. All predicted instance embeddings of the first frame are directly regarded as identified instances and saved into the memory bank. After processing all frames, our method produces a set of instances sequence. The majority votes are utilized to decide the unique category label of each instance sequence.
\section{Experiments}

\subsection{Dataset}
To verify the effectiveness of our approach, we evaluate it on recent three video instance segmentation benchmarks, YouTube-VIS-2019~\cite{yang2019video}, YouTube-VIS-2021~\cite{YouTube-VIS-2021} and OVIS-2021~\cite{qi2021occluded} datasets. Following previous works~\cite{yang2019video,bertasius2020classifying,yang2021crossover}, we evaluate our method on the validation sets of YouTube-VIS-2019, YouTube-VIS-2021 and OVIS-2021. 

\noindent \textbf{YouTube-VIS-2019} dataset contains 40 class annotations, including many common objects. The official dataset consists of three subsets: 2238 training videos, 302 validation videos, and 343 test videos.

\noindent \textbf{YouTube-VIS-2021} dataset is an improved version of YouTube-VIS-2019 containing 40 class annotations. It collects more videos and high-quality annotations. This dataset also consists of three subsets: 2985 training videos, 421 validation videos, and 453 test videos. 

\noindent \textbf{OVIS-2021} is a new large scale benchmark dataset for video instance segmentation task with 25 common semantic categories. It is designed with object occlusions in videos, which could reveal the complexity of real-world scenes. It consists of 607 training videos, 140 validation videos, and 154 testing videos as the official split.

\subsection{Metrics}
The evaluation metrics are average precision (AP) and average recall (AR), with the video Intersection over Union (IoU) of the mask sequences as the threshold~\cite{yang2019video}. Specifically, for a predicted mask $\hat{m}^i$ and a ground-truth mask $m^j$, we first extend them to the whole video with length T by padding empty mask. Then, 
\begin{equation}
    \text{IoU}(i,j) = \frac{\sum_{t=1}^T \hat{m}^i_t \cap m^j_t}{\sum_{t=1}^T \hat{m}^i_t \cup m^j_t }.
\end{equation}
According to the definition, if the model detects object masks successfully but fails to associate the objects across frames, it still gets a low IoU. Thus, accurate and robust instance association across frames is very crucial for achieving high performance.

\subsection{Implementation Details}\label{implement}
\noindent \textbf{Model Settings.} In our experiments, we choose the ResNet-50~\cite{he2016deep} and ResNet-101 with FPN~\cite{lin2017feature} as the backbone in the proposed method. Our model is pretrained on COCO $\mathtt{train2017}$ \cite{lin2014microsoft} with $1 \times$ schedule following previous works~\cite{cao2020sipmask,yang2021crossover,yang2019video}. We implement the proposed method with PyTorch~\cite{paszke2019pytorch} and the FPS is measured on an RTX $2080$ Ti GPU including the pre- and post-processing steps for fair comparison following previous work~\cite{yang2021crossover}. The optimizer of the proposed method is SGD, with a learning rate 5e-3 and a weight decay 1e-4. The models are trained with $1 \times$ schedule for 12 epoch, and we decay the lr with the ratio 0.1, in the 8-th and 11-th epoch. 
The input frames are resized to 640$\times$360 following previous works~\cite{yang2019video,yang2021crossover,fu2020compfeat}. 

\noindent \textbf{Hyper-parameters.} There exists some hyper-parameters in our proposed framework, \emph{i.e.}, bi-directional contrastive learning loss $\lambda_b$, and temporal consistency loss $\lambda_c$. In this paper, we set $\lambda_b = 0.2$ and $\lambda_c = 10$ in default.

\subsection{Main Results}\label{result}
Here we compare our method with two-stage~\cite{yang2018efficient,voigtlaender2019feelvos,yang2019video,bertasius2020classifying,fu2020compfeat}, single-stage~\cite{cao2020sipmask,wu2021track,liu2021sg}, 3D convolution-based~\cite{athar2020stem}, feature aggregation-based~\cite{li2021spatial}, and transformer-based~\cite{wang2021end} methods. For some differences in the training settings (\emph{e.g.}, resolution, training epochs) vary from different methods, we strictly follow MaskTrack R-CNN~\cite{yang2019video}, SipMask~\cite{cao2020sipmask} and CrossVIS~\cite{yang2021crossover} with $1 \times$ schedule and 640$\times$360 resolution for fair comparison.

\begin{table}[t!]
\begin{minipage}{0.46\linewidth}
\caption{Comparisons with some recent VIS methods on the \textbf{YouTube-VIS-2021} val set. We use ResNet-50 backbone and $1\times$ schedule for all experiments for fair comparison.}
\resizebox{\textwidth}{!}{
\begin{tabular}{l|ccccc}
\hlineB{2}
Methods & AP & AP$_{50}$ & AP$_{75}$ & AR$_{1}$ & AR$_{10}$ \\
  \hline
SipMask~\cite{cao2020sipmask} & 28.6 & 48.9 & 29.6 & 26.5 & 33.8\\
MaskTrack R-CNN~\cite{yang2019video} & 31.7 & 52.5 & 34.0 & 30.8 & 37.8\\
STEm-Seg~\cite{athar2020stem} & 33.3 & 53.8 & 37.0 & 30.1 & 37.6\\
CrossVIS~\cite{yang2021crossover} & 34.2 & 54.4 & 37.9 & 30.4 & 38.2\\
\bfseries Ours (STC) & \textbf{35.5} & \textbf{57.4} & \textbf{38.0} & \textbf{32.8} & \textbf{42.2}\\ 
\hlineB{2}
\end{tabular}}
\label{ytvis2021}
\end{minipage}
\quad
\begin{minipage}{0.46\linewidth}
\caption{Comparisons with some recent VIS methods on  very challenging \textbf{OVIS-2021} val set. We use ResNet-50 backbone and $1\times$ schedule for all experiments for fair comparison.}
\resizebox{\textwidth}{!}{
\begin{tabular}{l|ccccc}
\hlineB{2}
Methods & AP & AP$_{50}$ & AP$_{75}$ & AR$_{1}$ & AR$_{10}$ \\
  \hline
SipMask~\cite{cao2020sipmask} & 10.3 & 25.4 & 7.8 & 7.9 & 15.8\\
MaskTrack R-CNN~\cite{yang2019video} & 10.9 & 26.0 & 8.1 & 8.3 & 15.2\\
STEm-Seg~\cite{athar2020stem} & 13.8 & 32.1 & 11.9 & 9.1 & 20.0\\
CrossVIS~\cite{yang2021crossover} & 14.9 & 32.7 & 12.1 & 10.3 & 19.8\\
\bfseries Ours (STC) &  \textcolor{black}{\textbf{15.5}} & \textcolor{black}{\textbf{33.5}} & \textcolor{black}{\textbf{13.4}} & \textcolor{black}{\textbf{11.0}} & \textcolor{black}{\textbf{20.8}}\\ 
\hlineB{2}
\end{tabular}}
\label{ovis}
\end{minipage}
\end{table}

\noindent \textbf{YouTube-VIS-2019.} Without any bells and whistles, our proposed method achieves the best performance 36.7\% AP among the listed single-model methods. More specifically, among the two-stage methods, our model outperforms the original MaskTrack R-CNN~\cite{yang2019video} by \textbf{6.4} \% in AP (36.7\% vs. 30.3\%). As discussed in VisTR~\cite{wang2021end}, we also argue that the performance of MaskProp~\cite{bertasius2020classifying} relies heavily on stacking multiple networks, \emph{e.g.}, Spatio-temporal Sampling Network~\cite{bertasius2018object} and Hybrid Task Cascade Network~\cite{chen2019hybrid}, not to mention the larger resolution and more training epochs. Our model also beats the recently proposed CompFeat~\cite{fu2020compfeat} by 1.4 \% in AP with a significant improvement on the performance of speed. Meanwhile, it outperforms STEm-Seg~\cite{athar2020stem} and VisTR~\cite{wang2021end} with the same backbone on the accuracy, which indicates the superiority of our method. Note that VisTR utilizes multi-scale training and takes a week on 8 NVIDIA Tesla V100 for training. Furthermore, compared with the single-stage methods SipMask~\cite{cao2020sipmask} and TraDeS~\cite{wu2021track}, our method obtains about 4.2 \% and 4.1 \% improvement in AP, respectively. Compared with the feature aggregation-based method STMask~\cite{li2021spatial} which uses multi-frames to obtain more robust features, our method surpasses it by 3.2 \% in AP for ResNet-50 backbone, and even it uses a stronger ResNet-50-DCN backbone. When compared with recent work CrossVIS~\cite{yang2021crossover}, our method still shows the superiority of the performance on both performance and speed.

As shown in Table~\ref{sota}, we also compare the FPS (frames per second) with other state-of-the-art methods. Our method achieves 36.7\% AP at a 40.3 FPS, which is the best tradeoff for the single model. In addition, our method can run an online mode which is crucial for practical usages.

\noindent \textbf{YouTube-VIS-2021.} We evaluate the recently proposed MaskTrack R-CNN~\cite{yang2019video}, SipMask~\cite{cao2020sipmask} and CrossVIS~\cite{yang2021crossover} on YouTube-VIS-2021 using the official implementation for comparison. As shown in Table~\ref{ytvis2021}, our method surpasses MaskTrack R-CNN~\cite{yang2019video} and CrossVIS~\cite{yang2021crossover} by 3.8 \% and 1.3 \% in AP , which verifies the effectiveness of our method.

\noindent \textbf{OVIS-2021.} From Table~\ref{ovis} we can observe that all methods meet a large performance degradation due to the complexity and occlusions in OVIS-2021 dataset. Our method achieves the best 15.5\% AP, surpassing all methods under the same experimental conditions. We hope that our proposed method can serve as a strong baseline for this challenging benchmark.

\begin{table}[t!]
\begin{minipage}{0.46\linewidth}
\caption{Ablation studies for each component of the proposed framework on YouTube-VIS-2019 validation set.}
\centering
\resizebox{\textwidth}{!}{
\begin{tabular}{cccc|c}
\hlineB{2}
Baseline & Consistency & Contrastive & Bi-direction   & AP \\
  \hline
  \cmark& & &  &33.7 \\
  \cmark& \cmark & &  & 34.4\\
  \cmark & \cmark & \cmark& & 36.3\\
  \cmark & \cmark & \cmark & \cmark & \textbf{36.7}\\ 
\hlineB{2}
\end{tabular}}
\label{modules}
\end{minipage}
\quad
\begin{minipage}{0.46\linewidth}
\caption{Comparisons among different settings of the track embedding on the YouTube-VIS-2019 validation set.}
\centering
\resizebox{\textwidth}{!}{
\begin{tabular}{cc|c|c}
\hlineB{2}
Contrastive & Bi-direction & Embedding dim  & AP \\
  \hline
 \xmark & \xmark & 256 & 34.5  \\
 \cmark & \xmark & 256 &36.2   \\
 \xmark & \cmark & 256  &35.4     \\ 
 \cmark & \cmark & 256  & \textbf{36.7}\\ 
\hlineB{2}
\end{tabular}}
\label{tracking}
\end{minipage}
\end{table}

\subsection{Ablation Studies}\label{ablation}
We conduct experiments on the YouTube-VIS-2019 validation set with the ResNet-50 backbone and $1 \times$ schedule for the ablation studies.

\myparagraph{Analysis for Each Component.} As shown in Table~\ref{modules}, we first use CondInst~\cite{tian2020conditional} to obtain the instance masks instead of utilizing RoIAlign and mask head in MaskTrack R-CNN~\cite{yang2019video}, which achieves 3.4 \% in AP improvements (33.7\% vs. 30.3\%). Besides the performance improvement, this component also changes the two-stage model to a simple single-stage and fully convolutional one with faster speed. Note that our temporal consistency constraint for the kernel generator successfully gains 0.7 \% in AP by digging deeper into the temporal information in the video sequence. For the instances association across frames, we conduct experiments to verify the effectiveness of two components (``Contrastive" and ``Bi-direction"). Specifically, when only using spatio-temporal contrastive learning module, we could achieve 1.9\% in AP improvement. When using the bi-directional contrastive learning strategy, we finally obtain 36.7\% in AP, surpassing ``Contrastive" baseline by 0.4\% in AP, demonstrating the effectiveness of the bi-directional learning strategy.

\begin{table}[t!]
\begin{minipage}{0.45\linewidth}
\caption{Comparisons among different settings of the kernel generator head on the YouTube-VIS-2019 validation set.}
\centering
\begin{tabular}{c|c|c}
\hlineB{2}
Consistency& \# Conv & AP \\
  \hline
 \xmark & 1 &  31.5\\
 \xmark & 2 &  33.7\\
 \xmark & 3 & 36.2\\
 \xmark & 4 &  36.0   \\ \hline
 \cmark & 3 & \textbf{36.7}\\ 
\hlineB{2}
\end{tabular}
\label{dynamic_kernel}
\end{minipage}
\quad 
\begin{minipage}{0.45\linewidth}
\caption{Comparisons among different settings of the mask branch on the YouTube-VIS-2019 validation set.}
\centering
\begin{tabular}{c|c|c}
 \hlineB{2}
Coord. & \# Channel & AP \\
  \hline
 \xmark & 1 &  28.7\\
 \xmark & 4 &  33.6\\
 \xmark & 8 &  36.2 \\
 \xmark & 16 & 36.1  \\ \hline
\cmark & 8   &\textbf{36.7}\\ 
\hlineB{2}
\end{tabular}
\label{mask_base}
\end{minipage}
\end{table}

\begin{table}[t!]
\caption{Comparisons among different settings of the negative sampling methods of contrastive learning on the YouTube-VIS-2019 validation set.}
\centering
\setlength{\tabcolsep}{20 pt}
\begin{tabular}{c|c|c}
\hlineB{2}
Inbox& \# Negative  & AP \\
  \hline
\xmark & 0 & 34.9\\
 \xmark & 64 & 36.5 \\
\xmark & 128 & \textbf{36.7} \\
\xmark & 256 & 36.4
\\ \hline
\cmark & 128  & 35.2\\ 
\hlineB{2}
\end{tabular}
\label{negative}
\end{table}

\myparagraph{Kernel Generator.} Kernel generator from CondInst~\cite{tian2020conditional} plays a critical role in our method. Thus, we conduct ablation studies to show the impact of parameters in kernel generator head. As presented in Table~\ref{dynamic_kernel}, with the number of convolutions in kernel generator head increasing, the performance improves steadily and achieves the peak 36.7\% AP with three stacked convolutions. Temporal consistency obtains 0.5\% in AP, which demonstrates the effectiveness.

\myparagraph{Mask Branch.} To enhance the expressiveness of the mask feature, we further explore the channel number and relative coordinate map (``Coord.'') used in the mask branch. As illustrated in Table~\ref{mask_base}, the 8-channel mask feature achieve 36.2\% AP without the coordinate map, and extra channels cannot improve the performance. We set the channel number of the mask feature to 8 by default as a result. Relative coordinates are attached to the mask feature for better performance (about 0.5\% in AP improvement).

\myparagraph{Tracking Embedding.} The tracking embedding is crucial for VIS since AP relies heavily on the accuracy of instance association. We compare with different tracking embedding dimensions. As shown in Table~\ref{tracking}, AP improves as the embedding dimension increases. However, we can not afford the complexity cost when the embedding dimension is larger than 256 considering the speed. Thus, we set the embedding dimension as 256 by default.

\myparagraph{Negative Sampling.} The designed contrastive learning strategy aims to obtain representative and discriminative tracking embeddings. Thus, we further explore different numbers of negative embeddings and how they are selected in Table~\ref{negative}. ``Inbox'' means we randomly select the negative embeddings within boxes from negative locations according to label assignment results. We find that choosing 128 negative embeddings is a good balance of total training time and accuracy. Moreover, randomly selecting negative embeddings from the whole feature map of the reference frame is much better than ``Inbox''. This observation verifies that the model can learn more discriminative representations from background stuff or objects. 

\begin{figure}[t!]
\centering
\includegraphics[width=0.49\textwidth]{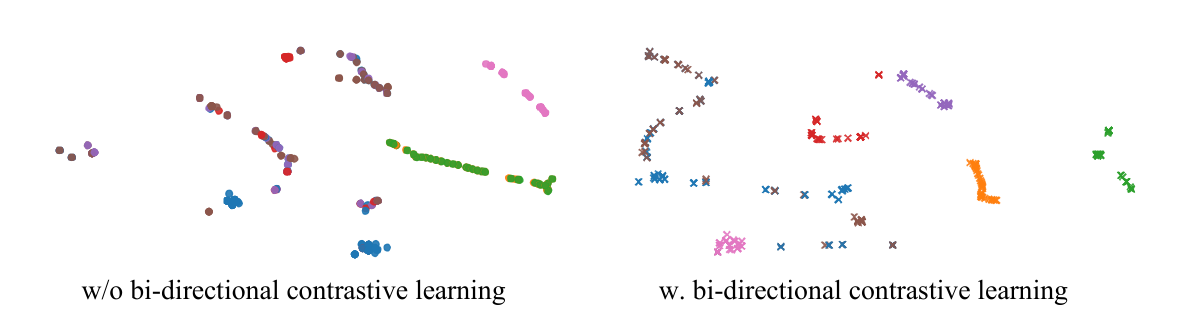}
\caption{Visualizations of instance embeddings without or with bi-directional contrastive learning module using t-SNE.}
\label{embeddings}
\end{figure}

\begin{figure*}[t!]
\centering
\includegraphics[width=0.92\textwidth]{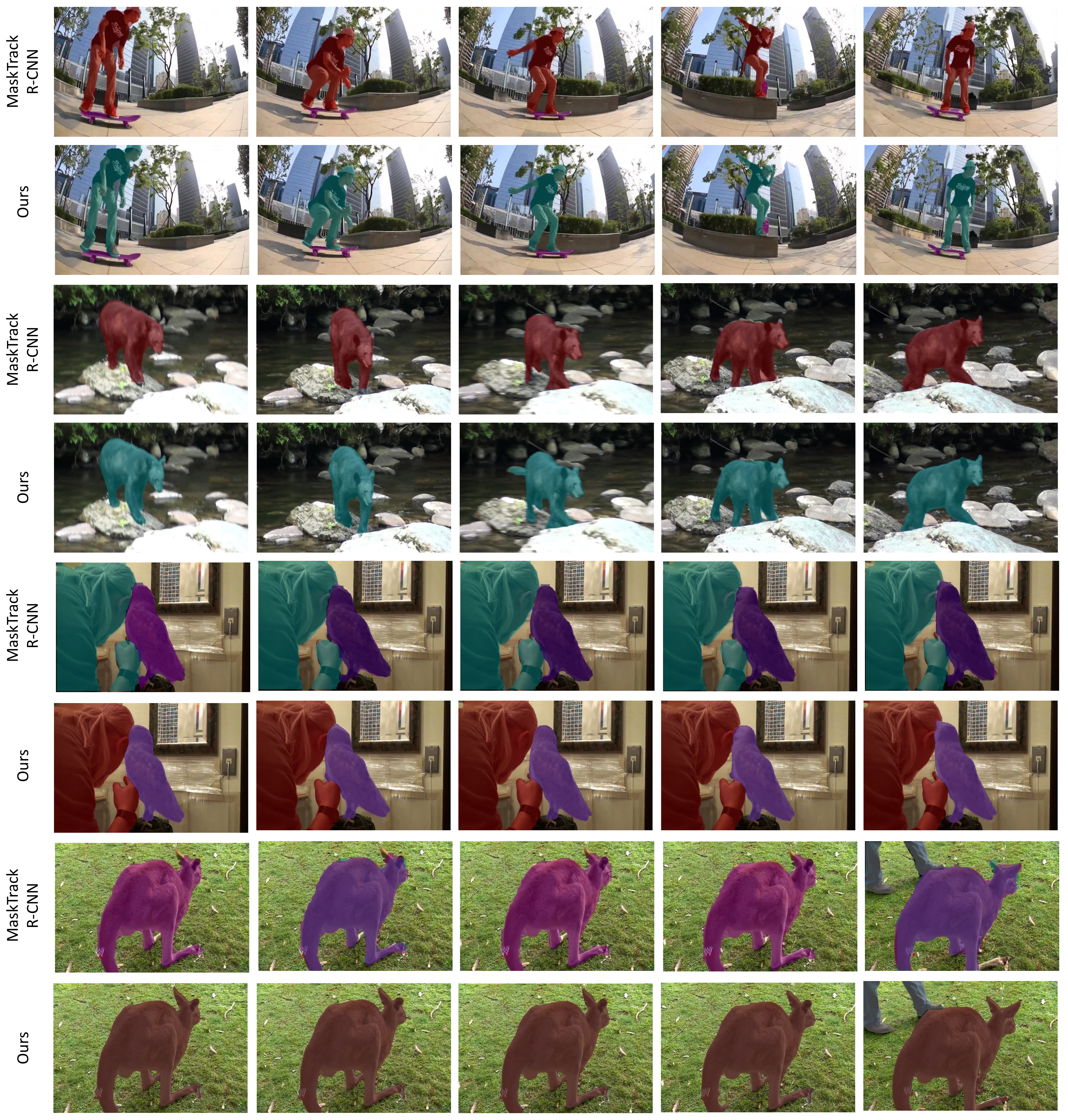}
\caption{Visualization of our proposed method and MaskTrack R-CNN on the YouTube-VIS-2019 val set.}
\label{visualizations}
\end{figure*}

\subsection{Visualizations}\label{sec:visualizations}
\myparagraph{Instance Embedding.} To verify the effectiveness of the proposed method qualitatively, we visualize the instance embeddings of the same video sequence using t-SNE~\cite{van2008visualizing}, which is shown in Figure~\ref{embeddings}. Comparing with Figure~\ref{embeddings}(a), the instance embeddings of Figure~\ref{embeddings}(b) is more separable, which indicates that our proposed STC module helps to distinguish different instances in the embedding space. Thus, compared with the original multi-class classification loss~\cite{yang2019video}, we could obtain more accurate instance association accuracy for video instance segmentation task.

\myparagraph{Video Visualization.} The visualization of the proposed method on the YouTube-VIS-2019 validation dataset is shown in Figure~\ref{visualizations}. Compared with baseline method MaskTrack R-CNN~\cite{yang2019video}, as shown in the first row and the second row, STC achieve more accurate segmentation results. From the last two rows, STC could achieve more coherent tracking results compared with MaskTrack R-CNN baseline, which demonstrates the effectiveness of the proposed spatio-temporal contrastive learning strategy. In conclusion, our method can segment and associate instances better with more accurate boundary results in challenging situations while MaskTrack R-CNN suffers from the missing instances or identity mistakes.
\section{Conclusion}
In this work, we introduced a effective architecture for video instance segmentation. Our model is conceptually simple without requiring RoIAlign operation or 3D convolutions. Moreover, it achieves state-of-the-art single-model results ({\em i.e.}, ResNet-50 backbone) on the YouTube-VIS-2019, YouTube-VIS-2021, and OVIS-2021 datasets in a fully convolutional fashion. We hope our work could serve an strong baseline, which could inspire designing more efficient framework and rethinking the embeddings loss for challenging video instance segmentation task.

\clearpage
%
%
\bibliographystyle{splncs04}
\bibliography{egbib}
\end{document}